\documentclass[10pt,twocolumn,letterpaper]{article}
\pdftrailerid{}
\usepackage{cvpr}
\newcommand{\methodname}{PrismCodec}
\newcommand{\sg}{\operatorname{sg}}
\newcommand{\E}{\mathbb{E}}
\newcommand{\R}{\mathcal{R}}
\newcommand{\zc}{\mathbf{z}_{c}}
\newcommand{\zclean}{\mathbf{z}_{0}}
\newcommand{\xev}{\widetilde{\mathbf{x}}}
\newcommand{\xrec}{\widehat{\mathbf{x}}}
\newcommand{\pstr}{\mathbf{P}^{\mathrm{str}}}
\newcommand{\psem}{\mathbf{P}^{\mathrm{sem}}}
\newcommand{\priors}{\mathcal{P}}
\usepackage[T1]{fontenc}
\usepackage{float}
\usepackage{tabularx}
\newcolumntype{Y}{>{\raggedright\arraybackslash}X}
\usepackage{flafter}
\usepackage{microtype}
\definecolor{cvprblue}{rgb}{0.21,0.49,0.74}
\usepackage[pagebackref,breaklinks,colorlinks,allcolors=cvprblue]{hyperref}
\hypersetup{pdftitle={Recasting DiT for Extreme Image Compression with Hierarchical Visual Priors},pdfauthor={Yunuo Chen, Jingnan Gao, Bing He, Jinjing Zhao, Chuqin Zhou, Xiaoyue Ling, Li Song, Guo Lu, Wenjun Zhang},pdfcreator={},pdfproducer={}}
\DeclareRobustCommand{\appref}[1]{%
  \hyperref[#1]{Appendix~\ref*{#1}}}

\title{Recasting DiT for Extreme Image Compression with Hierarchical Visual Priors}
\author{%
Yunuo Chen\textsuperscript{1}\quad
Jingnan Gao\textsuperscript{1}\quad
Bing He\textsuperscript{1}\quad
Jinjing Zhao\textsuperscript{2}\quad
Chuqin Zhou\textsuperscript{1}\\
Xiaoyue Ling\textsuperscript{1}\quad
Li Song\textsuperscript{1}\quad
Guo Lu\textsuperscript{1}\thanks{Corresponding author.}\quad
Wenjun Zhang\textsuperscript{1}\\[0.5em]
\textsuperscript{1}Shanghai Jiao Tong University\quad
\textsuperscript{2}University of Sydney
}

\makeatletter
\appto\@maketitle{%
  \begingroup
  \setlength{\columnwidth}{\textwidth}%
  \begin{figure}[H]
    \centering
    \includegraphics[width=\linewidth]{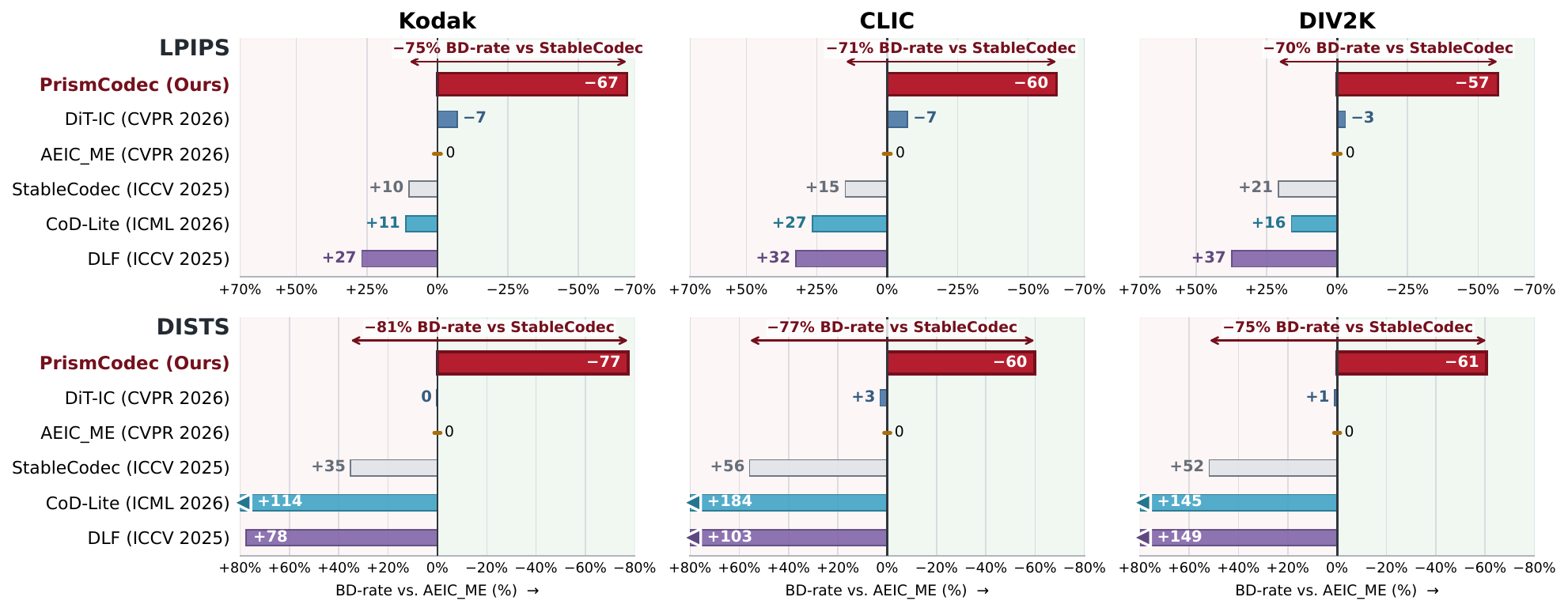}
    \vspace{-15pt}
    \caption{\textbf{Perceptual bitrate savings.} LPIPS- and DISTS-based BD-rate comparisons on Kodak, CLIC, and DIV2K, with AEIC-ME as the bar reference ($0\%$). Negative values indicate bitrate savings and extend to the right. Red arrows report PrismCodec's direct BD-rate relative to StableCodec. Triangles mark bars clipped at the axis limit; labels retain the unclipped values, rounded to integers.}
    \label{fig:teaser}
\end{figure}%
  \endgroup
}
\makeatother

\begin{document}
\maketitle
\begin{abstract}

Pretrained diffusion models provide powerful generative priors for extreme image compression, but their stochastic, iterative generation processes are not designed to decode compressed representations directly. We present PrismCodec, which recasts a pretrained Diffusion Transformer (DiT) as a bitstream-grounded one-step generative decoder. PrismCodec uses the decoded representation in two complementary roles: a projected codec state defines the starting state for generation, while hierarchical visual priors derived from the same decoded latent provide semantic and structural guidance. To bridge the gap between stochastic, iterative generation and deterministic one-step decoding, we develop a progressive adaptation strategy. We first learn codec-conditioned transport from a noisy neighborhood centered on the codec state, distill it into a one-step mapping, and finally anneal the source perturbation to zero during end-to-end optimization. The resulting decoder reconstructs deterministically from the received representation without additional conditioning bits. Across Kodak, CLIC, and DIV2K, PrismCodec consistently improves rate--perception performance over prior generative codecs, achieving BD-rate reductions of $70$--$75\%$ in LPIPS and $75$--$81\%$ in DISTS over StableCodec. Code will be released upon acceptance.
\par

\end{abstract}

\section{Introduction}
\label{sec:introduction}

At ultra-low bitrates, learned codecs built on optimized transforms and entropy models \citep{balle2016end,balle2018variational,minnen2018joint} can preserve only limited visual evidence of the source image. Generative compression therefore relies on image generation priors to synthesize details that are not explicitly transmitted. Yet compression differs fundamentally from unconstrained generation: plausible content is insufficient if it is inconsistent with the evidence retained in the bitstream. A central challenge is therefore to exploit powerful generative priors while grounding reconstruction in this limited transmitted evidence.

Pretrained diffusion and flow models offer powerful visual priors \citep{DDPM,LDM,dit,esser2024scaling}, but their generative processes are not naturally aligned with compression decoding. Existing methods expose image-specific compressed evidence to the generative decoder through different interfaces. OneDC~\citep{aigc_onedc} and DiT-IC~\citep{shi2026dit} derive semantic guidance directly from compressed representations, while AEIC~\citep{zhang2026ultra} removes the text-conditioning pathway and reconstructs from codec latents. NeFIC~\citep{chen2026nefic} and StableCodec~\citep{aigc_stablecodec} retain fixed text conditions but also inject image-specific information through a decoded anchor and a codec latent, respectively. Together, these approaches show that effective generative decoding depends on how transmitted visual evidence is exposed to the generator. Two questions nevertheless remain: \emph{how should that evidence determine both the generative starting state and the visual guidance, and how can a pretrained model built for stochastic, iterative generation be adapted to deterministic one-step decoding?}

We introduce \textbf{\methodname}, which recasts a pretrained DiT as a compression decoder whose starting state and visual guidance are both derived from the bitstream. These two roles are complementary: the codec-centered state anchors generation in the transmitted representation, while the visual priors specify semantic content and spatial structure that should be preserved during reconstruction. Inspired by recent image-generation studies \citep{shi2026latent,gong2026twins}, we combine semantic and structural visual priors through a shared visual-evidence pathway. A lightweight decoder renders the decoded codec features as an internal evidence image, providing a common image-domain interface through which pretrained visual encoders can interpret the same bitstream-derived evidence. DINOv3~\citep{simeoni2025dinov3} supplies multi-depth dense semantic features, which are delivered to successive DiT block groups through depth-matched cross-attention rather than collapsed into a single condition. This preserves complementary levels of visual abstraction across the pretrained representation hierarchy. In parallel, the diffusion-native VAE supplies a spatially aligned structural prior through the latent-input pathway. Both priors are computed entirely at the decoder and therefore require no additional conditioning bits.

Visual conditioning alone, however, does not make a pretrained generator a practical compression decoder. Diffusion and flow models are typically trained to generate from stochastic sources through iterative transport, whereas practical compression typically favors deterministic reconstruction from a transmitted representation under a tight decoding budget. We therefore progressively reshape the generative process around the codec state. First, we learn conditional transport from a stochastic neighborhood centered on the codec representation to the clean generative latent, so that the learned trajectory is already grounded in transmitted evidence rather than generic noise. We then distill this codec-conditioned transport into a direct one-step mapping. Finally, during end-to-end compression training, we gradually contract the stochastic source distribution onto the deterministic codec state. This annealing avoids an abrupt shift from the stochastic source used for flow learning and distillation to the deterministic input used at deployment, while allowing the codec and generative decoder to adapt jointly. Fig.~\ref{fig:teaser} summarizes the resulting rate--perception performance. Across Kodak, CLIC, and DIV2K, \methodname{} achieves LPIPS and DISTS BD-rate reductions of approximately $70$--$75\%$ and $75$--$81\%$, respectively, over StableCodec. Our main contributions are:

\begin{itemize}
    \setlength{\itemsep}{2pt}
    \setlength{\parsep}{0pt}
    \item We recast a pretrained DiT as a bitstream-grounded generative decoder, using the compressed representation to determine both its starting state and its visual guidance.
    \item We introduce hierarchical visual priors that combine multi-depth dense semantic features with a diffusion-native structural prior through complementary DiT conditioning interfaces.
    \item We develop a progressive adaptation strategy that moves from codec-centered stochastic flow learning through one-step distillation to deterministic end-to-end decoding, bridging pretrained stochastic generation and practical extreme image compression.
\end{itemize}

\section{Related Work}
\label{sec:related}

\textbf{Generative Image Compression.} Generative codecs use learned priors to recover plausible content at very low bitrates. Adversarial methods improve perceptual quality through distribution matching \citep{aigc_hific,aigc_MSILLM}, while conditional and residual diffusion reconstruct images from compressed representations \citep{aigc_cdc,aigc_DIRAC}. Recent work increasingly adapts pretrained generative priors, making bitstream grounding and efficient decoding central design concerns.

\begin{figure*}[t]
\centering
\includegraphics[width=\linewidth]{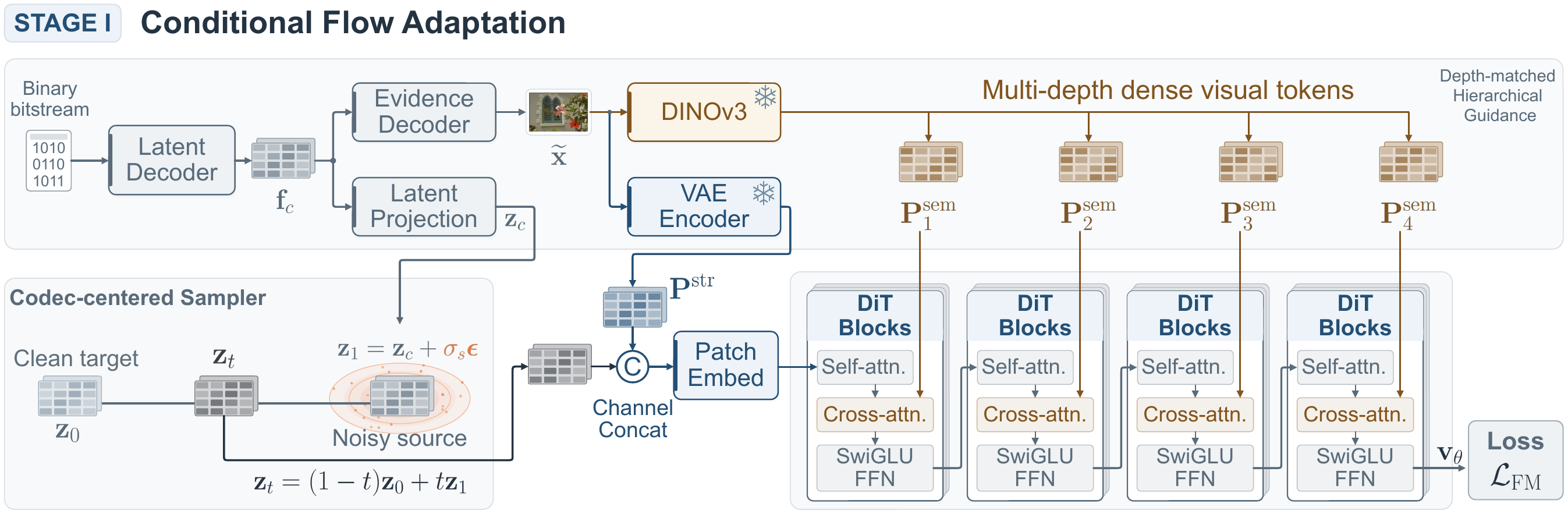}
\vspace{-13pt}
\caption{\textbf{Stage I: visual conditioning and codec-centered flow adaptation.} The decoded feature $\mathbf{f}_{c}$ yields a codec state $\zc$ and an evidence image $\xev$ for two frozen encoders. DINOv3 supplies multi-depth dense semantic priors $\psem$ to successive DiT groups through depth-matched cross-attention. The VAE encoder supplies a spatially aligned structural prior $\pstr$, channel-concatenated with $\mathbf{z}_{t}$. The codec-centered sampler interpolates between the clean latent and a noisy source centered on $\zc$. }
\label{fig:pipeline}
\vspace{-10pt}
\end{figure*}

\textbf{Bitstream Grounding and One-Step Adaptation.} Pretrained-prior codecs expose transmitted evidence through different conditioning interfaces to guide reconstruction \citep{aigc_stablecodec,chen2026nefic,chen2026steering}. StableCodec combines a codec latent with structural correction under fixed text conditioning. OneDC~\citep{aigc_onedc} derives tokenizer-supervised semantic tokens from the hyperprior, DiT-IC \citep{shi2026dit} uses text-aligned compressed features, and AEIC \citep{zhang2026ultra} removes cross-attention. PrismCodec instead separates bitstream-derived evidence into two roles: a codec state determines the generative source, while an evidence image provides hierarchical semantic and diffusion-native structural guidance through complementary pretrained encoders and separate DiT conditioning interfaces.

Fast decoding is addressed through diffusion pretraining and distillation. CoD~\citep{jia2026cod} pretrains diffusion for compression, CoD-Lite \citep{jia2026cod_lite} introduces lightweight one-step decoding, OSCAR \citep{guo2025oscar} supports one-step reconstruction across multiple bitrates, and FlowCodec~\citep{huang2026flowcodec} adapts a pretrained flow prior for near-terminal transport. More detailed comparisons with prior methods are provided in \appref{app:related_work_discussion}.
\par

\vspace{-5pt}
\section{Method}
\vspace{-5pt}
\label{sec:method}

\subsection{Overview}
\label{sec:method_overview}

PrismCodec recasts a pretrained DiT as a one-step generative decoder whose reconstruction is grounded in the received bitstream. Given a source image $\mathbf{x}$ encoded into a  bitstream $\mathbf{b}$, entropy decoding produces compressed features from which we derive two complementary forms of information: a codec state $\zc$ that determines the starting state for generation, and visual priors $\priors$ that specify the semantic content and spatial structure to preserve during reconstruction. The decoder computes
\begin{equation}
    \widehat{\mathbf{z}}_{0}
    =\mathcal{G}_{\theta}(\zc;\priors),
    \qquad
    \xrec=\mathcal{D}_{g}(\widehat{\mathbf{z}}_{0}),
    \label{eq:overview_final}
\end{equation}
where $\mathcal{G}_{\theta}$ is a one-step latent mapping implemented by a DiT, and $\mathcal{D}_{g}$ maps the predicted latent back to image space through the generative VAE decoder, kept frozen throughout all stages.

\begin{figure*}[!t]
\centering
\includegraphics[width=\linewidth]{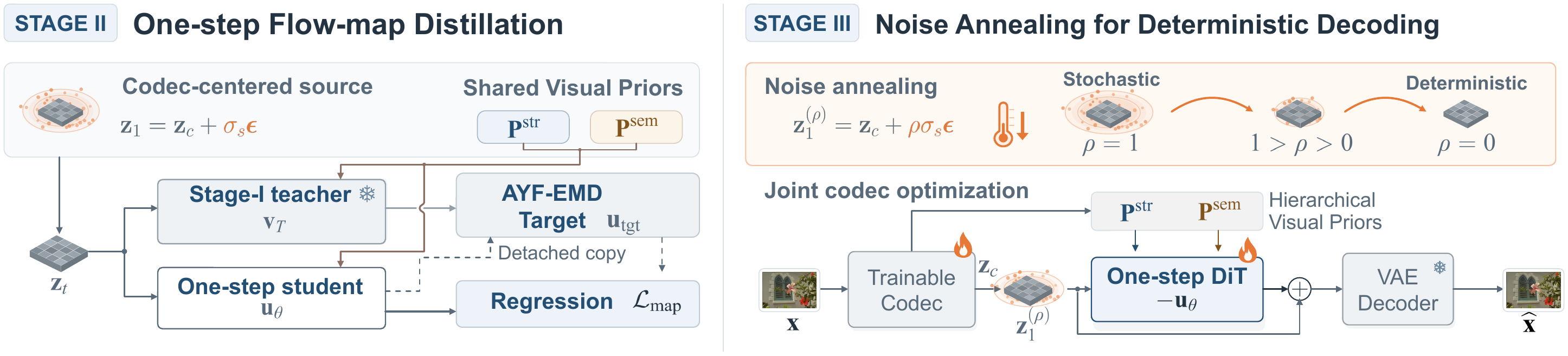}
\vspace{-15pt}
\caption{\textbf{Stages II and III: one-step distillation and deterministic source adaptation.} \emph{Left:} The frozen Stage-I teacher and student share paired interpolants $\mathbf{z}_t$ and visual priors $\priors$; separate conditioning arrows show the priors entering both models. The teacher and a detached copy of the current student form the AYF-EMD target $\mathbf{u}_{\mathrm{tgt}}$ for regression $\mathcal{L}_{\mathrm{map}}$, with destination time $r=0$ fixed throughout distillation. \emph{Right:} End-to-end optimization anneals $\rho:1\to0$, contracting the stochastic source onto the deterministic codec state $\zc$. Flames mark the jointly trained codec and DiT; the VAE decoder remains frozen. The residual sum combines $\mathbf{z}_{1}^{(\rho)}$ with the negative DiT prediction $-\mathbf{u}_{\theta}$. }
\label{fig:distillation_annealing}
\vspace{-10pt}
\end{figure*}

A pretrained diffusion model, however, is designed for stochastic, iterative generation rather than deterministic decoding from a transmitted representation. We therefore adapt the generative process progressively. Stage~I learns conditional transport from a stochastic neighborhood centered on the codec state, grounding the generative trajectory in transmitted evidence. Stage~II distills this codec-conditioned transport into a one-step map. Stage~III then contracts the remaining source randomness while jointly optimizing the codec and generator, continuously aligning the training-time source with the deterministic codec state used at deployment. Fig.~\ref{fig:pipeline} illustrates the visual interface and Stage-I adaptation, while Fig.~\ref{fig:distillation_annealing} summarizes Stages~II and III. \appref{app:notation} summarizes the notation.

\subsection{Bitstream-Derived Visual Evidence}
\label{sec:visual_evidence}

The decoded bitstream provides the common source for both generation initialization and visual guidance. Entropy decoding and synthesis recover a feature $\mathbf{f}_{c}$, from which PrismCodec constructs the codec state $\zc$ and an internal evidence image $\xev$. These two branches expose the same transmitted evidence through different interfaces: $\zc$ determines where the generative process starts, whereas $\xev$ provides an image-domain view from which pretrained visual representations can be extracted.

\noindent\textbf{Codec state.} A learned projection $\mathcal{P}_{c}$ maps the decoded feature to
\begin{equation}
    \zc=\mathcal{P}_{c}(\mathbf{f}_{c}).
\end{equation}

The projection places $\zc$ in coordinates compatible with the generative VAE latent in spatial dimensions and numerical scale, without constraining it to a Gaussian distribution. In Stage~I, $\zc$ centers the stochastic source; after Stage~III, it becomes the deterministic input to the one-step decoder. We denote the estimated bits per pixel (bpp) by $\R$.

\noindent\textbf{Shared visual interface.} In parallel, a lightweight evidence decoder $\mathcal{D}_{e}$ renders the same decoded feature into an image-domain representation,
\begin{equation}
    \xev=\mathcal{D}_{e}(\mathbf{f}_{c}).
    \label{eq:overview_evidence}
\end{equation}
Because $\xev$ is computed entirely from the received bitstream, it requires no additional transmission. More importantly, it provides a common image-domain interface through which the pretrained visual encoders interpret the same decoded evidence in the domain for which they were originally trained. We lightly supervise $\xev$ with an auxiliary objective $\mathcal{L}_{\mathrm{ev}}$ so that it retains source layout, appearance, and semantic content. The objective combines pixel reconstruction with DINOv3 feature similarity. Together, $\zc$ and $\xev$ expose the same decoded representation through the two roles required by the generative decoder: source initialization and visual guidance.

\subsection{Hierarchical Visual Priors}
\label{sec:visual_priors}

The evidence image is interpreted through two complementary pretrained representation spaces. Multi-depth DINOv3 features provide dense semantic guidance, while the diffusion-native VAE provides a reference in the generator's native latent space \citep{shi2026latent,gong2026twins}. Both priors are derived from $\xev$ at the decoder and therefore introduce no additional conditioning bits.

\noindent\textbf{Multi-depth dense semantic prior.} We extract spatial patch tokens from $L$ selected depths of a frozen DINOv3 transformer, excluding class and register tokens. After per-depth normalization and learned projection, they form
\begin{equation}
    \psem=\mathcal{H}(\xev)
    =\bigl(\mathbf{P}^{\mathrm{sem}}_{1},\ldots,
           \mathbf{P}^{\mathrm{sem}}_{L}\bigr),
    \label{eq:overview_priors}
\end{equation}
where each $\mathbf{P}^{\mathrm{sem}}_{\ell}$ is a dense spatial token sequence projected to the DiT hidden dimension. Rather than collapsing the selected DINOv3 depths into a single condition, we preserve their ordering and assign successive semantic levels to successive groups of DiT blocks through cross-attention. This depth-matched design exposes the generator to complementary levels of visual abstraction throughout the network.

\noindent\textbf{Diffusion-native structural prior.} The semantic branch does not directly provide a reference in the native latent coordinates used by the generator. We therefore encode the same evidence image with the frozen native VAE encoder. Let $\mathcal{S}$ denote the encoder with its fixed latent calibration. We define
\begin{equation}
    \pstr=\mathcal{S}(\xev),\qquad
    \zclean=\mathcal{S}(\mathbf{x}),
    \label{eq:structural_prior}
\end{equation}
where $\pstr$ is the structural prior and $\zclean$ is the clean latent used as the training target. Because $\pstr$ lies in the same latent coordinates as the generative state, it provides spatially aligned guidance complementary to the semantic tokens. Its role is distinct from that of $\zc$: the codec state defines the source from which transport begins, whereas $\pstr$ acts as a structural reference throughout the generative mapping.

\noindent\textbf{Complementary conditioning.} The complete visual condition is $\priors=(\pstr,\psem)$. The two priors enter the DiT through distinct interfaces: $\pstr$ is concatenated with the evolving latent before patch embedding, whereas the assigned semantic tokens are injected through cross-attention. We denote the resulting predictor by $\mathbf{u}_{\theta}(\mathbf{z},t,r;\priors)$, where $t$ and $r$ denote the current and destination flow times, respectively. Both conditioning paths remain active throughout adaptation and deployment.

\subsection{Learning a One-Step Generative Decoder}
\label{sec:progressive_training}

The pretrained generator and the deployed codec differ along two coupled axes: the former starts from a stochastic source and follows iterative transport, whereas the latter must start from a bitstream-determined state and decode in one deterministic step. Rather than imposing both changes at once, we adapt the generative process progressively. Stage~I re-centers the source on the codec state while retaining stochasticity for flow learning; Stage~II distills the resulting codec-conditioned transport into a one-step map; and Stage~III contracts the remaining source randomness while jointly adapting the complete compression system to its deployed deterministic input.

\subsubsection{Stage I: Conditional Flow Adaptation}
\label{sec:stage1}

We first ground the generative source in the transmitted representation while retaining stochasticity to facilitate flow learning. Specifically, we define a codec-centered source
\begin{equation}
    \mathbf{z}_{1}=\zc+\sigma_s\boldsymbol{\epsilon},
    \qquad
    \boldsymbol{\epsilon}\sim\mathcal{N}(\mathbf{0},\mathbf{I}),
    \label{eq:conditional_source}
\end{equation}
where $\sigma_s>0$ controls the source-noise scale. Generation therefore starts from a stochastic neighborhood around $\zc$, rather than from zero-centered noise. This follows the general intuition that an informative reconstruction can serve as the center of a generative source \citep{ohayon2025pmrf}.

We pair this source with the clean latent $\zclean$ through the linear interpolation
\begin{equation}
    \mathbf{z}_{t}=(1-t)\zclean+t\mathbf{z}_{1},
    \qquad t\sim\mathcal{U}(0,1),
    \label{eq:stage1_path}
\end{equation}
and train the DiT velocity predictor $\mathbf{v}_{\theta}(\mathbf{z}_{t},t;\priors)$ with
\begin{equation}
    \begin{aligned}
    \mathcal{L}_{\mathrm{FM}}
    &=\E\!\left[
      \left\|\mathbf{v}_{\theta}(\mathbf{z}_{t},t;\priors)
      -(\mathbf{z}_{1}-\zclean)\right\|_{2}^{2}
    \right],\\
    \mathcal{L}_{\mathrm{I}}
    &=\mathcal{L}_{\mathrm{FM}}
     +\lambda_{\mathrm{rate}}^{\mathrm{I}}\R
     +\lambda_{\mathrm{ev}}^{\mathrm{I}}\mathcal{L}_{\mathrm{ev}}.
    \end{aligned}
    \label{eq:stage1_total}
\end{equation}
We jointly optimize the codec, latent projection, evidence decoder, semantic projections, and DiT, while keeping the pretrained DINOv3 and VAE encoders frozen. The resulting model defines an iterative conditional transport from the codec-centered source to the clean latent and serves as the teacher for Stage~II. Sampling uses the learned velocity directly without CFG.

\subsubsection{Stage II: One-Step Flow-Map Distillation}
\label{sec:stage2}

Stage~I grounds the generative trajectory in the codec state, but the learned transport remains iterative. We therefore distill it into a direct map from the same source distribution to the clean endpoint. Importantly, the student distills the codec-conditioned transport learned in Stage~I, rather than the original pretrained generative trajectory. We adopt AYF's Eulerian map distillation (AYF-EMD) \citep{sabour2025align}. For distillation, the student predictor $\mathbf{u}_{\theta}(\mathbf{z},t,r;\priors)$ estimates the average velocity over the interval $[r,t]$ for $r<t$. At $r=t$, the same parameterization represents the instantaneous velocity used in Stage~I:
\[
    \mathbf{v}_{\theta}(\mathbf{z},t;\priors)
    \equiv
    \mathbf{u}_{\theta}(\mathbf{z},t,t;\priors).
\]
The Stage-I model is frozen as the teacher $\mathbf{v}_{T}$, while the student is initialized from its weights. The codec and all condition-producing modules are also frozen during this stage.

Distillation reuses the codec-centered source and paired interpolation in Eqs.~\eqref{eq:conditional_source}--\eqref{eq:stage1_path}, with the same visual priors $\priors$ shared by teacher and student. Following the average-velocity parameterization \citep{geng2025meanflow}, we fix the destination time to $r=0$ and train the student to represent transport directly to the clean endpoint. At $(t,r)=(1,0)$, the resulting one-step decoder is
\begin{equation}
    \mathcal{G}_{\theta}(\mathbf{z};\priors)
    =\mathbf{z}-\mathbf{u}_{\theta}(\mathbf{z},1,0;\priors).
    \label{eq:one_step_map}
\end{equation}

The student is trained with the AYF-EMD regression objective
\begin{equation}
    \mathcal{L}_{\mathrm{II}}
    =\mathcal{L}_{\mathrm{map}}
    =\E\!\left[
      w(t,0)
      \left\|
      \mathbf{u}_{\theta}(\mathbf{z}_t,t,0;\priors)
      -\sg(\mathbf{u}_{\mathrm{tgt}})
      \right\|_2^2
    \right],
    \label{eq:stage2_total}
\end{equation}
where $w(t,0)\geq0$ is the regression weight, $\mathbf{u}_{\mathrm{tgt}}$ is the AYF-EMD target constructed from the frozen teacher and a detached copy of the current student, and $\sg$ denotes stop-gradient. Only the student is updated in Stage~II, so the codec state and visual conditions remain fixed throughout distillation.

\subsubsection{Stage III: Noise Annealing for Deterministic Decoding}
\label{sec:stage3}

The distilled student already provides one-step decoding, but its source remains stochastic because Stage~II is trained from $\zc+\sigma_s\boldsymbol{\epsilon}$. Deployment, in contrast, should depend only on the received bitstream. We remove this train--deployment mismatch during end-to-end optimization by gradually annealing the source perturbation:
\begin{equation}
    \mathbf{z}_{1}^{(\rho)}
    =\zc+\rho\sigma_s\boldsymbol{\epsilon},
    \qquad
    \widehat{\mathbf{z}}_0
    =\mathcal{G}_{\theta}(\mathbf{z}_{1}^{(\rho)};\priors),
    \label{eq:source_contraction}
\end{equation}
where $\rho$ decreases from $1$ to $0$ over training. At $\rho=1$, the starting distribution matches that used for flow adaptation and distillation; as $\rho\rightarrow0$, it contracts to the codec state $\zc$. This gradual transition avoids replacing the stochastic distilled source with the deterministic deployment input abruptly, while allowing the codec and one-step decoder to adapt jointly along the contraction. The final decoder therefore recovers Eq.~\eqref{eq:overview_final} and requires no stochastic input at deployment.

During this stage, we jointly optimize the codec, evidence path, semantic projections, and one-step DiT; DINOv3 and both the VAE encoder and decoder remain frozen. The end-to-end objective is
\begin{equation}
    \mathcal{L}_{\mathrm{III}}
    =\lambda_{\mathrm{rate}}^{\mathrm{III}}\R
     +\mathcal{L}_{\mathrm{img}}
     +\lambda_{\mathrm{adv}}\mathcal{L}_{\mathrm{adv}}^{\mathrm{DINO}}
     +\lambda_{\mathrm{ev}}^{\mathrm{III}}\mathcal{L}_{\mathrm{ev}}.
    \label{eq:stage3_total}
\end{equation}
Here, $\mathcal{L}_{\mathrm{img}}$ combines MSE, MS-SSIM, LPIPS, and DISTS \citep{wang2003multiscale,lpips,dists}, while $\mathcal{L}_{\mathrm{adv}}^{\mathrm{DINO}}$ uses lightweight discriminator heads on frozen multi-depth DINOv3 features \citep{vfm_aid_gan_loss}. The evidence objective continues to regularize $\xev$ as the codec representation evolves during joint optimization. Construction and training details, loss definitions, and supporting analyses are provided in \appref{app:method_details}.

\begin{figure*}[!t]
\centering
\includegraphics[width=\linewidth]{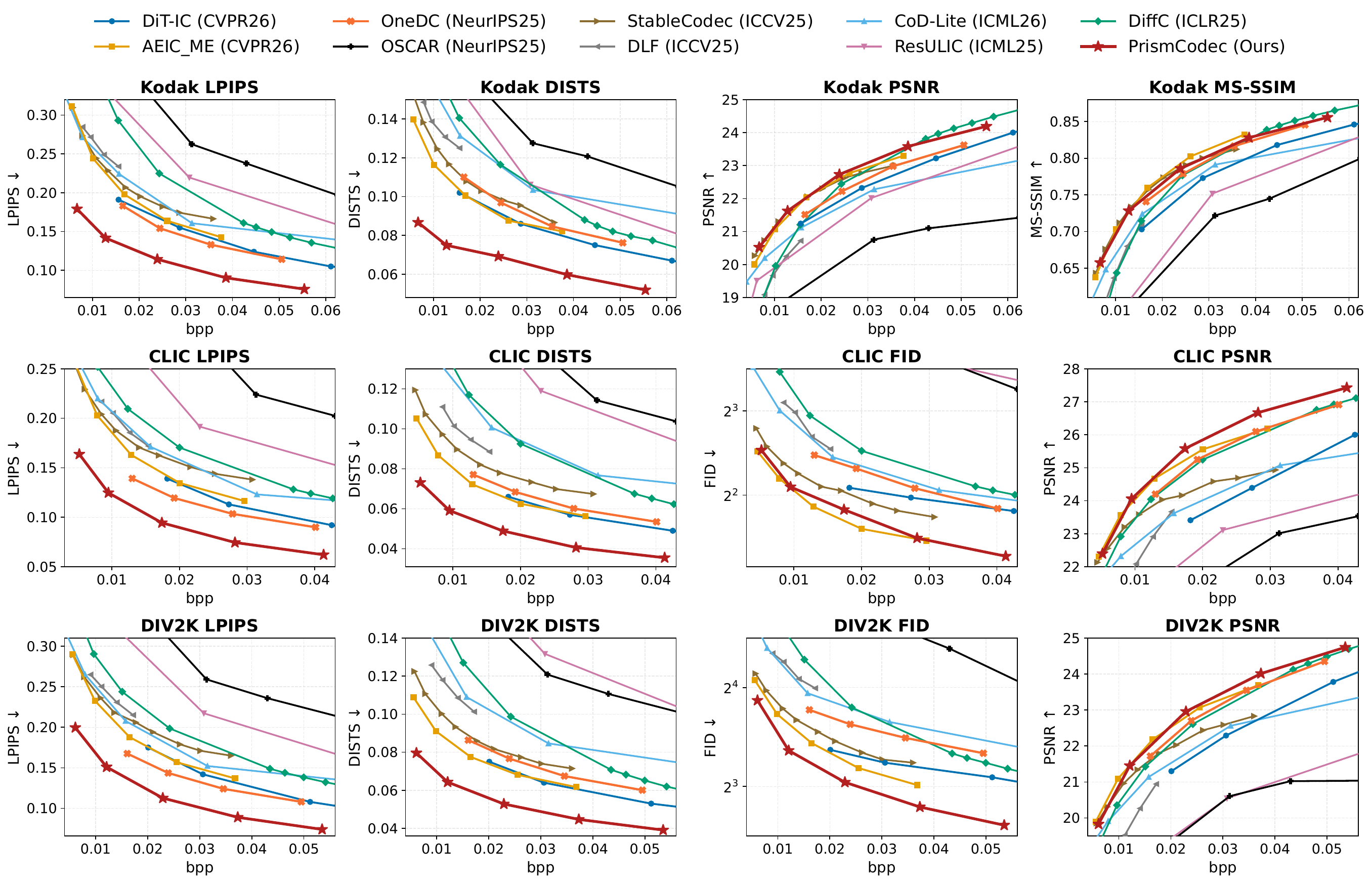}
\vspace{-20pt}
\caption{Rate--distortion and rate--perception comparisons on Kodak, CLIC2020, and DIV2K. }
\label{fig:main_rd}
\vspace{-10pt}
\end{figure*}

\section{Experiments}
\label{sec:experiments}

\subsection{Experimental Setup}
\label{sec:experimental_setup}

\subsubsection{Model and Training}
Following \citet{yao2025vavae}, we pretrain a class-conditional LightningDiT on ImageNet-1K~\citep{deng2009imagenet} for 100 epochs, and then add visual cross-attention layers for Stage~I adaptation. Following DiT-IC~\citep{shi2026dit}, we train on high-quality images from CLIC 2020 Professional~\citep{CLIC2020}, MLIC-Train-100K~\citep{jiang2023mlic++}, and LSDIR~\citep{li2023lsdir}, retaining images whose height and width both exceed 512 pixels. We use aspect-ratio buckets centered on a nominal $768\times768$ resolution, spanning ratios from $1{:}2$ to $2{:}1$. Before Stage~I, we pretrain the codec and evidence path for 20k updates using the evidence MSE and DINO losses to obtain a stable initialization, with the same weights as Stage~I: $\lambda_{\mathrm{ev}}^{\mathrm{I}}=1$ and $\lambda_{\mathrm{ev\mbox{-}sem}}=0.5$. We run Stages~I and II once at a high bitrate to obtain a one-step initialization. Stage~III first optimizes the highest-rate codec, followed by end-to-end fine-tuning toward lower rates with 50k updates per rate. All stages use AdamW on 16 A100 GPUs with a global batch size of 16. For the initial high-rate training, Stages~I and III each run for 50k updates with learning rates of $10^{-4}$ and $5\times10^{-5}$, respectively, while Stage~II runs for 20k updates at $5\times10^{-5}$. We train five models with $\lambda_{\mathrm{rate}}^{\mathrm{III}}\in\{1,2,4,8,16\}$ and set $\sigma_s=0.3$. For Stage~III, we use $\lambda_{\mathrm{mse}}=1.0$, $\lambda_{\mathrm{lpips}}=1.0$, $\lambda_{\mathrm{dists}}=0.5$, and $\lambda_{\mathrm{ms}}=0.1$ for the final-image objective, with $\lambda_{\mathrm{adv}}=0.5$ and $\lambda_{\mathrm{ev}}^{\mathrm{III}}=0.2$ under $\lambda_{\mathrm{ev\mbox{-}sem}}=0.5$. For the initial Stage~III run, $\rho$ follows a half-cosine decay from 1 to 0 over the first 10k updates and remains at 0 for the remaining 40k updates. We enable the adversarial loss after 30k Stage~III updates. For inference, large images are decoded with overlapping latent tiles corresponding to $1024\times1024$ pixels, with a 256-pixel overlap. The evidence image is resized to $512\times512$ for DINOv3.

\begin{table*}[!t]
\centering
\caption{Detailed comparisons of BD-rate (\%) relative to StableCodec. --: no overlapping range.}
\label{tab:main_bdrate}
\begingroup
\setlength{\tabcolsep}{2.2pt}
\renewcommand{\arraystretch}{1.16}
\scriptsize
\resizebox{\linewidth}{!}{%
\begin{tabular}{@{}l*{12}{r}@{}}
\toprule
Method & \multicolumn{4}{c}{Kodak} & \multicolumn{4}{c}{CLIC2020} & \multicolumn{4}{c}{DIV2K} \\
\cmidrule(lr){2-5}\cmidrule(lr){6-9}\cmidrule(l){10-13}
 & LPIPS & DISTS & PSNR & MS-SSIM & LPIPS & DISTS & FID & PSNR & LPIPS & DISTS & FID & PSNR \\
\midrule
DiffC \citep{aigc_diffc} & $84.91$ & $74.71$ & $34.58$ & $33.51$ & $48.49$ & $101.71$ & $181.51$ & $-7.28$ & $40.40$ & $101.89$ & $117.75$ & $4.03$ \\
ResULIC \citep{aigc_resulic} & $105.82$ & $76.51$ & $72.91$ & $91.57$ & $118.90$ & $274.69$ & $2117.98$ & $153.66$ & $113.46$ & $376.60$ & -- & $208.08$ \\
StableCodec \citep{aigc_stablecodec} & $0.00$ & $0.00$ & $0.00$ & $0.00$ & $0.00$ & $0.00$ & $0.00$ & $0.00$ & $0.00$ & $0.00$ & $0.00$ & $0.00$ \\
DLF \citep{aigc_addingdl} & $21.21$ & $29.35$ & $108.54$ & $57.23$ & $18.32$ & $42.11$ & $151.63$ & $75.98$ & $21.38$ & $70.13$ & $125.20$ & $81.32$ \\
OneDC \citep{aigc_onedc} & $-36.63$ & $-5.46$ & $23.88$ & $10.74$ & $-56.28$ & $-28.94$ & $87.88$ & $-34.31$ & $-52.46$ & $-9.98$ & $65.34$ & $-17.47$ \\
OSCAR \citep{guo2025oscar} & $237.66$ & $204.80$ & $308.30$ & $161.49$ & $369.67$ & $444.09$ & $713.43$ & $423.85$ & $256.87$ & $351.99$ & $470.60$ & $269.90$ \\
DiT-IC \citep{shi2026dit} & $-27.79$ & $-28.08$ & $40.25$ & $43.46$ & $-39.39$ & -- & $46.73$ & $51.02$ & $-31.57$ & $-32.62$ & $-13.07$ & $30.47$ \\
AEIC-ME \citep{zhang2026ultra} & $-9.10$ & $-26.03$ & $1.37$ & $-2.42$ & $-12.92$ & $-35.81$ & $-36.78$ & $-29.67$ & $-17.22$ & $-34.14$ & $-20.26$ & $-20.65$ \\
CoD-Lite \citep{jia2026cod_lite} & $0.97$ & $62.40$ & $55.68$ & $28.09$ & $10.35$ & $98.48$ & $107.12$ & $36.73$ & $-3.38$ & $77.42$ & $102.18$ & $23.37$ \\
\midrule
\textbf{\methodname} & $\mathbf{-75.49}$ & $\mathbf{-81.33}$ & $\mathbf{-5.82}$ & $\mathbf{0.40}$ & $\mathbf{-71.46}$ & $\mathbf{-77.08}$ & $\mathbf{-30.63}$ & $\mathbf{-30.45}$ & $\mathbf{-69.54}$ & $\mathbf{-74.64}$ & $\mathbf{-46.96}$ & $\mathbf{-17.21}$ \\
\bottomrule
\end{tabular}%
}
\endgroup
\par\vspace{2pt}
\begin{minipage}{\linewidth}
\scriptsize
\end{minipage}
\end{table*}

\begin{figure*}[!t]
\centering
\includegraphics[width=\linewidth]{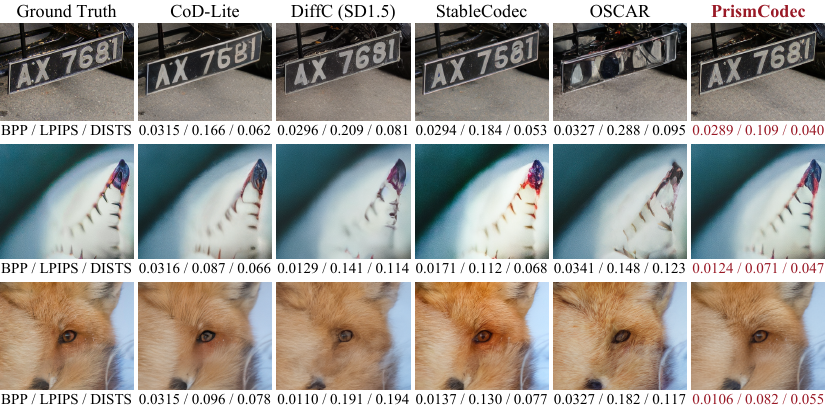}
\vspace{-20pt}
\caption{\textbf{Qualitative comparisons on DIV2K:} 0847, 0857, and 0862 (top to bottom). }
\label{fig:qualitative_comparison}
\vspace{-10pt}
\end{figure*}
\vspace{-3pt}
\subsubsection{Datasets, Metrics, and Baselines}
\vspace{-3pt}
We evaluate on Kodak \citep{kodak1993} (24 images), the CLIC2020 test set \citep{CLIC2020} (428 images), and the DIV2K validation set \citep{div2k} (100 images). We report PSNR and MS-SSIM \citep{wang2003multiscale} for distortion, LPIPS \citep{lpips} and DISTS \citep{dists} for perceptual fidelity, and FID \citep{FID} for distributional realism. The four paired metrics are computed on full-resolution RGB reconstructions and averaged over images; FID uses $256\times256$ patches on CLIC2020 and DIV2K, following \citet{aigc_hific}. We omit FID on Kodak given its small sample size. We compare with DiffC \citep{aigc_diffc}, ResULIC \citep{aigc_resulic}, DLF \citep{aigc_addingdl}, OSCAR \citep{guo2025oscar}, StableCodec \citep{aigc_stablecodec}, OneDC \citep{aigc_onedc}, AEIC-ME \citep{zhang2026ultra}, DiT-IC \citep{shi2026dit}, and CoD-Lite \citep{jia2026cod_lite}. We summarize bitrate efficiency with BD-rate \citep{bjontegaard2001}, using StableCodec as the reference.  Negative BD-rate indicates bitrate savings at equal quality.

\subsection{Main Results}
\label{sec:main_results}

\subsubsection{Quantitative Evaluation}
The main comparison tests whether grounding both the generative source and visual guidance in the received bitstream translates into a better rate--perception tradeoff without systematically sacrificing source fidelity. Fig.~\ref{fig:main_rd} reports the rate--distortion and rate--perception curves in the ultra-low-bitrate regime, while Table~\ref{tab:main_bdrate} summarizes BD-rates relative to StableCodec.

\paragraph{Perceptual fidelity and realism.}
Across all three datasets, \methodname{} shifts the LPIPS and DISTS curves favorably relative to prior generative codecs. On CLIC2020, it reduces bitrate relative to StableCodec by 71.46\% at matched LPIPS and 77.08\% at matched DISTS; the corresponding savings on DIV2K are 69.54\% and 74.64\%. The same operating points also yield FID-based savings of 30.63\% and 46.96\% on CLIC2020 and DIV2K, respectively, indicating that the gains in paired perceptual fidelity are accompanied by improved distributional realism. \appref{app:kid_evaluation} provides complementary KID comparisons.

\paragraph{Source fidelity.}
The perceptual gains do not come with a systematic loss in distortion performance. \methodname{} achieves PSNR-based bitrate savings of 5.82\%, 30.45\%, and 17.21\% on Kodak, CLIC2020, and DIV2K, respectively. MS-SSIM remains close to StableCodec, with a 0.40\% bitrate increase on Kodak and reductions of 1.48\% and 0.90\% on CLIC2020 and DIV2K. Taken together, these results support the central premise of PrismCodec: bitstream-grounded generation can improve perceptual reconstruction while retaining source-dependent fidelity.

\begin{table*}[!t]
\centering
\setlength{\abovecaptionskip}{0pt}
\setlength{\belowcaptionskip}{1pt}
\caption{Encoding and decoding efficiency at $2048\times2048$.}
\label{tab:efficiency}
\begingroup
\fontsize{8pt}{8pt}\selectfont
\setlength{\tabcolsep}{2pt}
\renewcommand{\arraystretch}{0.85}
\begin{tabular*}{\linewidth}{@{\extracolsep{\fill}}lrrrrrrrrr@{}}
\toprule
\raisebox{-6.5pt}[0pt][0pt]{\textbf{Method}} &
\raisebox{-6.5pt}[0pt][0pt]{\textbf{Params (B)}} &
\multicolumn{2}{c}{\textbf{Latency (s)}} &
\multicolumn{3}{c}{\textbf{kMACs/pixel}} &
\multicolumn{3}{c}{\textbf{FLOPs (T)}} \\ [-1.5pt]
\cmidrule(lr){3-4}\cmidrule(lr){5-7}\cmidrule(lr){8-10}
 & & Enc. & Dec. & Enc. & Dec. & Total & Enc. & Dec. & Total \\
\midrule
StableCodec & 1.07 & 4.65 & 9.74 & 2,965.0 & 10,343.7 & 13,308.7 & 24.87 & 86.77 & 111.64 \\
AEIC-ME & 0.93 & 0.08 & 5.16 & 204.0 & 5,959.2 & 6,163.2 & 1.71 & 49.99 & 51.70 \\
DiT-IC & 1.05 & 0.50 & 0.83 & 2,759.9 & 4,528.5 & 7,288.4 & 23.15 & 37.99 & 61.14 \\
DLF & 1.23 & 0.41 & 1.00 & 2,727.5 & 5,250.2 & 7,977.7 & 22.88 & 44.04 & 66.92 \\
OneDC & 1.40 & 0.65 & 1.51 & 2,888.1 & 10,292.4 & 13,180.5 & 24.23 & 86.34 & 110.57 \\
OSCAR & 1.05 & 0.89 & 3.22 & 3,150.1 & 10,890.4 & 14,040.5 & 26.42 & 91.36 & 117.78 \\
\midrule
\textbf{\methodname} & 1.89 & 0.62 & 2.10 & 1,352.7 & 13,571.8 & 14,924.5 & 11.35 & 113.85 & 125.20 \\
\bottomrule
\end{tabular*}
\par
\vspace{-10pt}
\endgroup
\end{table*}

\begin{table*}[!ht]
\centering
\begin{minipage}[t]{0.49\linewidth}
\vspace{0pt}
\begingroup
\centering
\fontsize{8pt}{8pt}\selectfont
\setlength{\abovecaptionskip}{0pt}
\setlength{\belowcaptionskip}{3pt}
\caption{\raggedright Ablations of visual priors.}
\label{tab:ablation_evidence_source}
\fontsize{8pt}{8pt}\selectfont
\setlength{\tabcolsep}{3pt}
\renewcommand{\arraystretch}{0.96}
\begin{tabularx}{\linewidth}{@{}Xrr@{}}
\toprule
\textbf{Variant} & \textbf{LPIPS (\%)} & \textbf{DISTS (\%)} \\
\midrule
Single semantic depth & $-59.64$ & $-68.37$ \\
Structural only & $-51.67$ & $-58.32$ \\
Semantic only & $-42.89$ & $-47.24$ \\
\midrule
\textbf{Full \methodname} & $\mathbf{-69.54}$ & $\mathbf{-74.64}$ \\
\bottomrule
\end{tabularx}
\vspace{-7pt}
\par
\endgroup
\end{minipage}\hfill
\begin{minipage}[t]{0.49\linewidth}
\vspace{0pt}
\begingroup
\centering
\fontsize{8pt}{8pt}\selectfont
\setlength{\abovecaptionskip}{0pt}
\setlength{\belowcaptionskip}{3pt}
\caption{\raggedright Ablations of source and training.}
\label{tab:ablation_training}
\fontsize{8pt}{8pt}\selectfont
\setlength{\tabcolsep}{3pt}
\renewcommand{\arraystretch}{0.96}
\begin{tabularx}{\linewidth}{@{}Xrr@{}}
\toprule
\textbf{Variant} & \textbf{LPIPS (\%)} & \textbf{DISTS (\%)} \\
\midrule
Zero-centered noise & $-37.81$ & $-46.80$ \\
w/o annealing & $-55.02$ & $-54.41$ \\
w/o distillation & $-54.74$ & $-58.04$ \\
\midrule
\textbf{Full \methodname} & $\mathbf{-69.54}$ & $\mathbf{-74.64}$ \\
\bottomrule
\end{tabularx}
\vspace{-7pt}
\par
\endgroup
\end{minipage}
\end{table*}

\subsubsection{Qualitative Evaluation}
\vspace{-3pt}
Fig.~\ref{fig:qualitative_comparison} examines whether the quantitative gains correspond to source-specific visual details rather than only aggregate metric improvements. At lower bitrates than the compared baselines, \methodname{} better preserves plate lettering (DIV2K 0847), leaf-tip contours and serrations (DIV2K 0857), and the fox's eye, nose, and fur patterns (DIV2K 0862), whereas several baselines smooth or alter these details. These examples are qualitatively consistent with the intended role of visual evidence in constraining generative reconstruction. \appref{app:qualitative} provides additional comparisons.

\vspace{-5pt}
\subsection{Efficiency Evaluation}
\vspace{-5pt}
\label{sec:efficiency}
We  report deployed parameters, encoding and decoding latency, and computation (kMACs/pixel and FLOPs) in Table~\ref{tab:efficiency}. Measurements use synthetic RGB inputs at $2048\times2048$, batch size one, and an NVIDIA A100 GPU; after warm-up, latency is averaged over 30 CUDA-synchronized runs. \methodname{} decodes a 2K image in 2.10\,s, compared with 9.74\,s for StableCodec and 3.22\,s for OSCAR, while remaining in the same order of total computation as these two generative baselines.

\subsection{Ablation Studies}
\vspace{-3pt}
\label{sec:ablations}
We organize the ablations around the three claims underlying PrismCodec: (i) semantic and structural visual priors provide complementary guidance, with multi-depth semantics contributing beyond a single representation level; (ii) transmitted evidence should determine the generative source as well as its conditions; and (iii) explicit one-step distillation and subsequent source contraction are both important for adapting stochastic generation to deterministic decoding. All results report LPIPS- and DISTS-based BD-rate (\%) on DIV2K relative to StableCodec. Each variant trains its own high-rate model and uses matched fine-tuning schedules to obtain the five rate points in Sec.~\ref{sec:experimental_setup}.

\paragraph{Hierarchical visual priors.}
Table~\ref{tab:ablation_evidence_source} separates two aspects of the proposed conditioning design. First, \emph{Single semantic depth} repeats the final selected DINOv3 layer across all four DiT groups, increasing LPIPS- and DISTS-based BD-rate by 9.90 and 6.27 percentage points relative to the full model. This isolates the benefit of preserving multiple representation depths rather than collapsing semantic guidance to one level. Second, \emph{Structural only} removes semantic guidance and increases the two BD-rates by 17.87 and 16.32 points, while \emph{Semantic only} removes structural input conditioning and increases them by 26.65 and 27.40 points. The degradation in both single-branch variants supports the complementary roles of dense semantic guidance and diffusion-native structural guidance.

\paragraph{Bitstream-grounded source.}
With both priors retained, \emph{Zero-centered noise} replaces the codec-centered source with a Gaussian-noise starting state and performs worse than both the stochastic codec-centered variant (\emph{w/o annealing}) and the full model (Table~\ref{tab:ablation_training}). Thus, the gain from $\zc$ is not merely duplicated by the conditioning paths: exposing the bitstream through the generative source provides an additional and complementary form of grounding.

\paragraph{Progressive adaptation to deterministic decoding.}
Table~\ref{tab:ablation_training} evaluates the two transitions from the Stage-I iterative flow to deterministic one-step decoding. \emph{w/o annealing} keeps $\rho=1$ through Stage~III and inference, using $\zc+0.3\boldsymbol{\epsilon}$ as the source, and increases LPIPS- and DISTS-based BD-rates by 14.52 and 20.23 points. This shows that the stochastic distilled source should be contracted to the deterministic codec state before deployment. \emph{w/o distillation} skips Stage~II and directly fine-tunes the Stage-I model with the full model's annealing schedule, increasing the two BD-rates by 14.80 and 16.60 points. Together, these results support both flow-map distillation and source-noise annealing in the transition to deterministic one-step decoding.
\par

\vspace{-3pt}
\section{Conclusion}
\vspace{-3pt}
\label{sec:conclusion}
We presented \methodname, which recasts a pretrained DiT as a bitstream-grounded one-step decoder for extreme image compression. The compressed representation determines both the generative starting state and hierarchical semantic and structural guidance. Through codec-centered flow adaptation, flow-map distillation, and source-noise annealing, \methodname progressively transforms stochastic iterative generation into deterministic one-step decoding. The resulting model reconstructs directly from the received bitstream without additional conditioning bits, yielding an efficient and effective rate--perception tradeoff.

\clearpage
\appendix
\section{Extended Related Work}
\label{app:related_work_discussion}

This appendix expands the positioning in Sec.~\ref{sec:related}. We first
place PrismCodec within learned and generative image compression, and then
compare the closest pretrained-prior codecs along the same two questions
that motivate our method: \emph{how bitstream-derived evidence enters the
generator} and \emph{how stochastic generation is adapted to one-step
reconstruction}.

\subsection{Conventional Learned Compression}

Learned codecs have advanced through stronger entropy models
\citep{minnen2020channel,he2021checkerboard,he2022elic,qian2022entroformer,jiang2023mlic,jiang2023mlic++,cheng2020learned,du2025largeLLM},
more expressive transforms
\citep{li2023frequency,liu2023learned,zeng2025mambaic,wu2025cmamba,chen2025s2cformerreorientinglearnedimage,chen2026contentaware_cmic,chen2026adaptive,li2026inevitable},
and specialized, scalable, or distilled designs for efficiency
\citep{aigc_add7,aigc_add9,aaai_reference,cvpr2025diction,iccv25HPCM,scaleup,chen2025knowledgeKDIC,li2025differentiableVMAF,zhang2023neural,liang2025structureGS}.
These approaches improve the coded representation itself; PrismCodec instead
focuses on how a pretrained generative prior should decode the limited
evidence that remains after extreme compression.

\subsection{Broader Generative Compression}

Generative codecs introduce learned priors to recover visually plausible
content at low rates. Existing families use adversarial priors
\citep{GAN,aigc_gan0,aigc_hific,aigc_MSILLM,aigc_MR,aigc_egic,aigc_add8},
discrete representations
\citep{vqvae,esser2021taming,aigc_add1,aigc_vqgan,aigc_vqgan1,aigc_glc},
or complementary representations
\citep{aigc_dual1,aigc_addingdl,zhou2026dual,ling2026free}; related work also studies
bitrate control, progressive decoding, and rate--distortion-optimized
quantization
\citep{li2024once,zhang2026autoregressive,cao2026progic,jiang2026differentiable}.

Diffusion-based codecs reconstruct from compressed information through
conditional denoising
\citep{aigc_add4,aigc_cdc,aigc_HFD}, residual generation
\citep{aigc_DIRAC,aigc_adding2}, or pretrained generative priors
\citep{podell2023sdxl,flux2024_dev,aigc_FDM,aigc_add3}. Other methods learn
latent guidance and reconstruction corrections
\citep{aigc_diffeic,aigc_corrdiff,aigc_diffpc,aigc_picd,aigc_resulic,aigc_icme},
or use frozen priors for zero-shot coding
\citep{aigc_diffc,aigc_addd,elata2024psc,ohayon2025compressed,vaisman2026turbo};
recent work further connects diffusion noise with quantization
\citep{relic2025bridging,du2025noise}. CoD pretrains diffusion specifically
for compression \citep{jia2026cod}, and CoD-Lite combines this with a
lightweight one-step convolutional decoder \citep{jia2026cod_lite}.
Accelerated sampling and distillation
\citep{ddim,vprediction,sauer2024adversarial,yin2024one,yin2024improved}
also motivate one-step codecs \citep{guo2025oscar,park2026single}, while
FlowCodec uses LoRA-adapted flow priors for near-terminal one-step latent
transport with an empty prompt \citep{huang2026flowcodec,hu2022lora}.

A second line of work concerns the conditioning interface. Text-guided
codecs transmit descriptions or learn language-aligned representations
\citep{aigc_text_sketch,aigc_perco,aigc_perco_image,korber2025percov2,aigc_taco,aigc_gan2,aigc_adding1},
and CADC derives text from reconstructions \citep{sheng2026cadc}. Image
conditions are also widely used for controlled generation and restoration
\citep{control1,control2,control3,chen2024faithdiff}, while REPA aligns
diffusion hidden states with pretrained visual features during training
\citep{yu2024representation}.

For the closest compression methods, Table~\ref{tab:conditioning_comparison}
summarizes how the transmitted representation determines the generator input
and conditions, together with the mechanism used for one-step adaptation.
The categories below concern the explicit conditioning pathway; all methods
also receive image-specific information from their coded representations.

\begin{table*}[!htbp]
\centering
\caption{\textbf{Comparison of bitstream grounding and one-step adaptation.}
The middle column summarizes how image-specific coded evidence enters the
generator; the right column summarizes how each method obtains fast or
one-step decoding. Fixed text conditions can coexist with image-specific
latent or anchor guidance.}
\label{tab:conditioning_comparison}
\small
\setlength{\tabcolsep}{4pt}
\renewcommand{\arraystretch}{1.12}
\begin{tabularx}{\linewidth}{@{}p{0.14\linewidth} Y Y@{}}
\toprule
Method & Generator input and conditioning & One-step adaptation\\
\midrule
StableCodec & Codec latent and structural residual branch; fixed text.
& SD-Turbo adaptation with bitrate pruning.\\
NeFIC & Anchor tokens and a codec-decoded bypass latent.
& Next-frame adaptation, then bypass training.\\
OneDC & Codec latent and tokenizer-supervised hyperprior tokens.
& One-step diffusion with semantic and perceptual distillation.\\
DiT-IC & Compressed latent, text-aligned context, and variance-derived
pseudo-timesteps.
& Reconstruction flow and encoder-latent self-alignment.\\
AEIC & Codec latent and structural residual; cross-attention removed.
& Pruned SD-Turbo decoder and feature distillation for shallow encoders.\\
SODEC & Transformed codec latent and ViT features of an auxiliary image.
& Fixed-timestep SD2.1 adaptation with rate annealing.\\
\methodname & Codec-centered source, multi-depth semantic tokens, and a
VAE structural condition.
& Conditional flow adaptation, flow-map distillation, and source contraction.\\
\bottomrule
\end{tabularx}
\end{table*}

\subsection{Fixed Text Conditions}

StableCodec~\citep{aigc_stablecodec} adapts SD-Turbo with a fixed prompt and
a codec-produced latent input. Its auxiliary encoder supplies source
features during compression, and an auxiliary decoder predicts a structural
residual that is added after denoising and before VAE decoding. The fixed
prompt therefore coexists with substantial image-specific guidance.
PrismCodec separates the same broad grounding problem into distinct roles:
a codec-centered source determines where generation starts, while a frozen
VAE latent and multi-depth visual tokens provide structural and semantic
guidance through separate DiT interfaces. StableCodec also tightens the rate
penalty during training; PrismCodec's source contraction instead changes the
generator's input distribution as it approaches deterministic deployment.

NeFIC~\citep{chen2026nefic} also retains a fixed prompt, using a decoded
anchor for conditioning and a codec-derived bypass latent as the input to
its video-diffusion generator.

\subsection{Learned Semantic Surrogates: OneDC and DiT-IC}

OneDC~\citep{aigc_onedc} decodes a semantic token sequence from the quantized
hyperprior and uses it as cross-attention context. A pretrained generative
tokenizer supplies training labels through an auxiliary prediction task;
the tokenizer and predictor are unnecessary at inference. The resulting
hyperprior tokens retain spatial locality. PrismCodec instead computes semantic conditions from an evidence image at
decoding time, preserves patch-token sets from multiple pretrained depths,
and assigns them to successive DiT block groups. This makes the semantic
hierarchy an explicit inference-time condition. OneDC additionally uses
diffusion distillation for perceptual refinement; its tokenizer supervision
and refinement therefore address different roles from PrismCodec's
Stage-II flow-map distillation.

DiT-IC~\citep{shi2026dit} projects compressed features into the text
conditioning space and uses contrastive alignment with text embeddings
during training; inference uses the latent condition alone. Its adaptation
also includes a variance-derived spatial pseudo-timestep and self-distillation
that aligns reconstructed latents with frozen encoder outputs.
PrismCodec preserves multiple visual representation depths and adds a
separate diffusion-native VAE structural condition. Its one-step adaptation
also follows a different path: Stage~I first learns continuous conditional
transport around the codec state, and Stage~II distills the resulting
finite-time map. The source perturbation scale therefore defines a Gaussian
neighborhood of the transmitted codec state, whereas DiT-IC's pseudo-timestep
controls spatially varying reconstruction strength.

\subsection{Removing the Cross-Attention Pathway: AEIC}

AEIC~\citep{zhang2026ultra} simplifies the SD-Turbo denoiser by removing the
text encoder, timestep embeddings, and cross-attention layers. Its codec
synthesis transform produces both a denoising input and a structural
residual, so reconstruction still depends on image-specific compressed
information. Dual-side feature distillation transfers knowledge from a
moderate-encoder model to shallow-encoder variants. This targets encoder
efficiency and feature preservation. PrismCodec retains explicit visual cross-attention and adapts a
time-dependent conditional flow before distilling its map in Stage~II.
The two uses of distillation therefore solve different problems: AEIC
targets encoder efficiency and feature transfer, whereas PrismCodec
compresses an already bitstream-grounded generative transport into one step.

\subsection{Auxiliary Reconstruction Guidance: SODEC}

SODEC~\citep{chen2026steering} is closely related to our use of a decoded
image as a visual interface. Its fidelity decoder reconstructs an auxiliary
image from compressed latents; a frozen ViT and a learned projection produce
a cross-attention condition. A separate transform supplies the
input to an SD2.1 U-Net evaluated at a fixed timestep. Both methods therefore derive visual guidance and the generative input from
the bitstream. PrismCodec further separates the conditioning role into two
pretrained representation spaces: semantic features remain distinct across
DINOv3 depths, while a diffusion-native VAE latent provides spatially aligned
structural guidance. The shared evidence image gives both encoders a common
image-domain view of the decoded evidence, and its auxiliary objective
includes paired semantic supervision (\appref{app:evidence}).

The adaptation mechanisms also differ. SODEC pretrains a high-bitrate
codec, warms up the diffusion path, and then tightens the rate penalty during
joint optimization. PrismCodec instead learns codec-centered conditional
transport, distills that transport into a one-step map, and then contracts
the stochastic source onto the deterministic codec state. SODEC's rate
annealing changes the coding constraint; PrismCodec's source-noise annealing
changes the generator's input distribution.

\subsection{Visual Priors and Progressive Flow Adaptation}

PrismCodec's semantic branch uses dense DINOv3 features
\citep{simeoni2025dinov3}, while its structural branch uses the frozen
VAE encoder. Both are explicit
inference-time conditions derived from the same bitstream-dependent evidence
image. This differs from representation-alignment methods such as
REPA~\citep{yu2024representation}, which use pretrained visual features as
training supervision for diffusion hidden states. More generally, using
semantic features in an auxiliary loss or discriminator is not equivalent
to supplying those features as decoder-side visual context.

The adaptation axis is likewise compositional. Stage~I builds on flow
matching and rectified flow \citep{lipman2023flowmatching,liu2022flow};
Stage~II uses established AYF-EMD distillation with an interval-average
velocity parameterization \citep{sabour2025align,geng2025meanflow}.
PMRF~\citep{ohayon2025pmrf} provides a related precedent for centering a
stochastic flow source on an informative deterministic estimate. PrismCodec
instead centers the source on an entropy-decoded codec state, without a
posterior-mean assumption or inherited PMRF optimality guarantee, and couples
that source with hierarchical visual conditions. Stage~III then specializes
the distilled map to the deterministic codec state used at deployment
(Sections~\ref{sec:stage1}--\ref{sec:stage3}).

\section{Notation}
\label{app:notation}

Tables~\ref{tab:notation} and~\ref{tab:notation_details} collect the symbols
that define the two central parts of PrismCodec: bitstream-derived visual grounding and progressive
adaptation to deterministic one-step decoding. Bold symbols denote
tensor-valued states, features, predictions, or ordered collections.
Superscripts $\mathrm{str}$ and $\mathrm{sem}$ identify the two visual-prior
types, while $\mathrm{I}$--$\mathrm{III}$ denote training stages. Latent
subscripts $0$ and $1$ denote flow endpoints. The Stage-III multiplier
$\rho$ controls source contraction independently of the flow times.

\begin{table*}[!ht]
\centering
\caption{Core notation used in the main text and pipeline figures.}
\label{tab:notation}
\small
\setlength{\tabcolsep}{5pt}
\renewcommand{\arraystretch}{1.1}
\begin{tabularx}{\linewidth}{@{}p{0.34\linewidth} Y@{}}
\toprule
Symbol & Meaning\\
\midrule
$\mathbf{x},\mathbf{b}$ & Source image and transmitted binary bitstream.\\
$\mathbf{f}_c,\zc$ & Shared decoded feature and projected codec-centered latent state.\\
$\xev$ & Internal evidence image produced by the visual-evidence decoder.\\
$\zclean$ & Clean target latent from the frozen VAE encoder.\\
$\mathbf{z},\mathbf{z}_t$ & Generic generative latent and paired training interpolant at time $t$.\\
$\mathbf{z}_1,\mathbf{z}_1^{(\rho)}$ & Perturbed codec states in Stages I/II and Stage III.\\
$\widehat{\mathbf{z}}_0,\xrec$ & Predicted generative latent and final reconstruction.\\
$\mathcal{P}_c,\mathcal{D}_e$ & Latent projection and visual-evidence decoder acting on $\mathbf{f}_c$.\\
$\mathcal{G}_{\theta},\mathcal{D}_g$ & One-step latent map and frozen, calibrated generative VAE decoder.\\
$\mathcal{S},\mathcal{H}$ & Structural mapping and hierarchical semantic mapping.\\
$\pstr$ & Structural prior: a VAE latent with the same shape as $\mathbf{z}$.\\
$\mathbf{P}^{\mathrm{sem}}_\ell,\psem$ & Semantic patch-token matrix at level $\ell$ and the ordered hierarchy of all $L$ levels.\\
$\priors$ & Joint visual condition $(\pstr,\psem)$, fixed along flow time.\\
$L,\ell$ & Number of semantic levels and level index $1\leq\ell\leq L$.\\
$t,r$ & Current and destination flow times; $r=t$ in Stage~I and $r=0$ in Stages~II/III.\\
$\boldsymbol\epsilon,\mathbf{0},\mathbf{I}$ & Standard Gaussian noise, zero vector, and identity of the latent dimension.\\
$\sigma_s$ & Per-coordinate source perturbation standard deviation.\\
$\rho$ & Scheduled Stage-III perturbation multiplier, reduced from $1$ to $0$.\\
$\mathbf{v}_{\theta},\mathbf{v}_{T}$ & Trainable instantaneous velocity and frozen teacher velocity.\\
$\mathbf{u}_{\theta},\theta$ & DiT predictor and its parameters; other module parameters are implicit.\\
$\R$ & Expected entropy-model rate estimate, in bits per pixel (bpp).\\
$\mathcal{L}_{\mathrm{FM}},\mathcal{L}_{\mathrm{map}}$ & Stage-I flow-matching loss and Stage-II regression loss.\\
$\mathcal{L}_{\mathrm{ev}},\mathcal{L}_{\mathrm{img}}$ & Paired evidence loss and final-image reconstruction loss.\\
$\mathcal{L}_{\mathrm{adv}}^{\mathrm{DINO}}$ & Generator-side adversarial loss on DINOv3 features.\\
$\mathcal{L}_{\mathrm{I}},\mathcal{L}_{\mathrm{II}},\mathcal{L}_{\mathrm{III}}$ & Total objectives of the three training stages.\\
$\lambda$ with subscripts & Non-negative weights of the named loss terms; stage superscripts distinguish their settings.\\
\bottomrule
\end{tabularx}
\end{table*}

\begin{table*}[!ht]
\centering
\caption{Additional notation for construction, distillation, and analysis.}
\label{tab:notation_details}
\small
\setlength{\tabcolsep}{5pt}
\renewcommand{\arraystretch}{1.1}
\begin{tabularx}{\linewidth}{@{}p{0.34\linewidth} Y@{}}
\toprule
Symbol & Meaning\\
\midrule
$\widehat{\mathbf{y}},\widehat{\mathbf{h}},p$ & Quantized main/hyperlatents and learned discrete probability masses.\\
$\E,\mathcal{N},\mathcal{U}$ & Expectation, Gaussian distribution (or its explicitly evaluated density), and uniform distribution.\\
$i_\ell,\Phi_{\mathrm{DINO}}^{(i_\ell)}$ & Zero-based DINOv3 block index and frozen feature extractor at that index.\\
$\operatorname{Patch},\operatorname{LN}_\ell,\mathcal{A}_\ell$ & Spatial-token selection, learned layer normalization, and learned per-depth MLP projection.\\
$N_\ell,n,\mathbf{F}_{\ell,n}$ & Patch count, patch index, and unit-normalized token for paired semantic supervision.\\
$\mathcal{L}_{\mathrm{ev}}^{\mathrm{DINO}}$ & Per-image paired DINOv3 feature term inside the expected evidence loss.\\
$K,k,\gamma(k)$ & DiT block count, one-based block index, and assigned semantic level.\\
$\mathbf{H}_0,\mathbf{H}_k$ & Patch-embedded input and output of DiT block $k$.\\
$\overline{\mathbf{H}}_k,\widetilde{\mathbf{H}}_k,\mathbf{e}_{t,r}$ & Intermediate block token sequences and time conditioning.\\
$\operatorname{SA},\operatorname{CA},\operatorname{FFN}$ & Self-attention, cross-attention, and feed-forward network.\\
$\operatorname{PatchEmbed},[\,\cdot\,;\,\cdot\,]$ & DiT patch embedding and ordered channel concatenation.\\
$\mathbf{v}_{\mathrm{opt}}$ & Conditional-mean velocity minimizing the fixed-construction flow loss.\\
$w(t,0)$ & Non-negative Stage-II regression weight for the interval $[0,t]$.\\
$\mathbf{u}_{\mathrm{tgt}},\sg$ & AYF-EMD regression target and stop-gradient operator.\\
$q_1,q_{1,\rho},\delta_{\zc}$ & Stage-I/II source density, Stage-III source law, and point mass at $\zc$.\\
$d,M,W_2$ & Latent coordinate count, fixed-decoder Lipschitz constant, and 2-Wasserstein distance.\\
$\mathcal{J},j,|\mathcal{J}|$ & Discriminator feature-index set, one of its indices, and its size.\\
$D_{\omega,j},\operatorname{Head}_{\omega,j},\omega$ & Feature-level discriminator, trainable head, and head parameters.\\
$\ell_D^+,\ell_D^-,\ell_G$ & Real-image, fake-image, and generator classification-loss functions.\\
$\mathcal{L}_D^{\mathrm{DINO}}$ & Discriminator-side objective, separate from the generator loss.\\
\bottomrule
\end{tabularx}
\end{table*}

\section{Additional Method Details}
\label{app:method_details}

\subsection{Codec State and Rate Estimate}
\label{app:codec_rate}

Entropy decoding and synthesis recover the shared feature $\mathbf{f}_c$,
from which PrismCodec exposes the transmitted evidence through two
complementary paths. The latent projection $\mathcal{P}_c$ produces
$\zc=\mathcal{P}_c(\mathbf{f}_c)$, which determines the generative starting
state, while the evidence decoder $\mathcal{D}_e$ produces
$\xev=\mathcal{D}_e(\mathbf{f}_c)$, the shared image-domain interface used
to construct the semantic and structural priors. Both paths are therefore
derived from the same received bitstream.

Let $\widehat{\mathbf{y}}$ and $\widehat{\mathbf{h}}$ denote the quantized
main latent and hyperlatent. For an RGB image of height $H$ and width $W$,
the expected rate estimate in bits per pixel (bpp) is
\begin{equation}
    \R=\E\!\left[
      \frac{
      -\log_2 p(\widehat{\mathbf{y}}\mid\widehat{\mathbf{h}})
      -\log_2 p(\widehat{\mathbf{h}})
      }{HW}
    \right].
    \label{eq:rate}
\end{equation}
Here, $p$ denotes discrete probability masses from the learned entropy
models, including the context factorization of the main latent. The
expectation averages training images and the codec's training-time
quantization, with each image's estimated code length normalized by its own
pixel count. Measured bpp uses the actual serialized stream length, including
coding overhead and required metadata, divided by the original image's pixel
count before averaging over images.

\subsection{Evidence-Preservation Objective}
\label{app:evidence}

Let $\mathbf{F}_{\ell}(\cdot)$ be the spatial patch-token matrix from the
$\ell$-th selected DINOv3 depth, with each token normalized to unit
$\ell_2$ norm. Write $N_{\ell}$ for its token count and
$\mathbf{F}_{\ell,n}$ for token $n$. The stop-gradient operator $\sg$
preserves its argument's value while blocking optimization gradients.
For an image pair, the semantic term is
\begin{equation}
    \mathcal{L}_{\mathrm{ev}}^{\mathrm{DINO}}
    =\frac{1}{L}\sum_{\ell=1}^{L}\frac{1}{N_{\ell}}
      \sum_{n=1}^{N_{\ell}}\left[
      1-\left\langle
        \mathbf{F}_{\ell,n}(\xev),
        \sg\bigl(\mathbf{F}_{\ell,n}(\mathbf{x})\bigr)
      \right\rangle\right].
    \label{eq:evidence_dino}
\end{equation}
The inner product is cosine similarity because the token features are
normalized. The source-image features are targets; their gradients are
stopped. DINOv3 parameters are frozen on both branches, while gradients
through its evidence-image input are retained to train $\mathcal{D}_{e}$
and, when enabled, the upstream codec. The total evidence loss averages
the paired terms:
\begin{equation}
    \mathcal{L}_{\mathrm{ev}}
    =\E\!\left[
      \operatorname{MSE}(\xev,\mathbf{x})
      +\lambda_{\mathrm{ev\text{-}sem}}
       \mathcal{L}_{\mathrm{ev}}^{\mathrm{DINO}}
    \right].
    \label{eq:evidence_total}
\end{equation}
Here, $\operatorname{MSE}$ averages squared differences over all pixels and
RGB channels, and $\lambda_{\mathrm{ev\text{-}sem}}$ balances paired pixel
fidelity and semantic consistency. Stages~I and III apply their respective
evidence-loss weights, as specified in the corresponding objectives.
Importantly, this objective supervises the \emph{interface} from which the
visual priors are extracted; it does not directly optimize the final
generative reconstruction.

\subsection{Prior Construction and Injection}
\label{app:injection}

The semantic branch preserves the depth structure of the pretrained visual
representation instead of collapsing it into one condition. Let $i_{\ell}$
be the zero-based DINOv3 block index selected for semantic level
$\ell\in\{1,\ldots,L\}$. The frozen feature extractor
$\Phi_{\mathrm{DINO}}^{(i_{\ell})}$ includes image preprocessing and the
applicable pretrained normalization; $\operatorname{Patch}$ retains only
spatial tokens. The projected semantic tokens are
\begin{equation}
    \mathbf{P}^{\mathrm{sem}}_{\ell}
    =\mathcal{A}_{\ell}\!\left(
      \operatorname{LN}_{\ell}\!\left(
        \operatorname{Patch}\!\left(
          \Phi_{\mathrm{DINO}}^{(i_{\ell})}(\xev)\right)\right)\right),
    \label{eq:semantic_adapter}
\end{equation}
where $\operatorname{LN}_{\ell}$ is a learned layer normalization and
$\mathcal{A}_{\ell}$ is a learned MLP projection into the DiT hidden
dimension. These operations preserve token order. The implementation uses
indices $(5,11,17,23)$, corresponding to the 6th, 12th, 18th, and 24th
transformer blocks. The last selected level uses DINOv3's pretrained final
normalization. Evidence crops are resized to $512\times512$ and normalized
with the visual encoder's input statistics; class and register tokens are
excluded. These operations are collectively denoted by $\mathcal{H}$ in
the main text.

The codec state $\zc$, structural prior $\pstr$, and clean target $\zclean$
share the frozen generative VAE's latent coordinates. The DiT concatenates the
evolving state with $\pstr$ along the channel dimension and predicts a
latent update with the same shape as the evolving state.

For the structural mapping $\mathcal{S}$, images in $[0,1]$ are converted to
the VAE input range $[-1,1]$. The encoder's posterior mode is calibrated by
subtracting the checkpoint's per-channel latent mean and dividing by its
per-channel standard deviation. $\mathcal{D}_g$ reverses this calibration
before VAE decoding and converts the decoded image back to $[0,1]$.
The fixed latent calibration is distinct from the source-noise scale
$\sigma_s$.

For $K$ DiT blocks divided into $L$ equal consecutive groups, with $K$
divisible by $L$, block $k\in\{1,\ldots,K\}$ receives semantic level
\begin{equation}
    \gamma(k)=1+\left\lfloor\frac{(k-1)L}{K}\right\rfloor,
    \label{eq:depth_mapping}
\end{equation}
where $\lfloor\cdot\rfloor$ is the floor operation. With $L=4$, the four
semantic levels guide four consecutive DiT groups, as shown in
Fig.~\ref{fig:pipeline}.

Let $\mathbf{H}_0=\operatorname{PatchEmbed}([\pstr;\mathbf{z}])$ be the
initial token sequence; $[\,\cdot\,;\,\cdot\,]$ concatenates channels in
the displayed order. $\mathbf{H}_k$ is the output of block $k$, so block $k$
receives $\mathbf{H}_{k-1}$. Let $\mathbf{e}_{t,r}$ denote its time
conditioning. Suppressing normalization and modulation details, the block
has the form
\begin{align}
    \overline{\mathbf{H}}_{k}
    &=\mathbf{H}_{k-1}
      +\operatorname{SA}_{k}(\mathbf{H}_{k-1};\mathbf{e}_{t,r}),
      \label{eq:prior_sa}\\
    \widetilde{\mathbf{H}}_{k}
    &=\overline{\mathbf{H}}_{k}
      +\operatorname{CA}_{k}
       \left(\overline{\mathbf{H}}_{k},
       \mathbf{P}^{\mathrm{sem}}_{\gamma(k)}\right),
      \label{eq:prior_ca}\\
    \mathbf{H}_{k}
    &=\widetilde{\mathbf{H}}_{k}
      +\operatorname{FFN}_{k}
       (\widetilde{\mathbf{H}}_{k};\mathbf{e}_{t,r}).
      \label{eq:prior_ffn}
\end{align}
Here, $\operatorname{SA}$, $\operatorname{CA}$, and $\operatorname{FFN}$ denote
self-attention, cross-attention, and the feed-forward network. The barred
and tilded token sequences are the intermediate outputs after self- and
cross-attention, respectively. The destination-time embedding is multiplied
by $t-r$ before combination with the current-time embedding; its contribution
therefore vanishes when $r=t$ in Stage~I. In Stage~II, $r=0$ is fixed and
only $t$ varies; Stage~III and deployment use $(t,r)=(1,0)$.
The structural prior is therefore present from the generator input,
whereas the semantic hierarchy supplies depth-dependent cross-attention
context throughout the DiT. This realizes the complementary roles described
in the main text without introducing any separately transmitted condition.

\subsection{Conditional Flow Targets}
\label{app:conditional_flow}

This subsection clarifies what Stage~I learns when the source is centered
on the codec state. Hold the codec and condition-producing parameters fixed,
and let the training construction induce a joint law of
$(\zclean,\mathbf{z}_1,\priors)$. For a fixed $t$ and
$\mathbf{z}_t=(1-t)\zclean+t\mathbf{z}_1$, least-squares regression has the
population optimum
\begin{equation}
    \mathbf{v}_{\mathrm{opt}}(\mathbf{z},t;\priors)
    =\E\!\left[\mathbf{z}_1-\zclean
      \mid \mathbf{z}_t=\mathbf{z},\priors\right].
    \label{eq:conditional_mean_velocity}
\end{equation}
This follows by conditioning the squared error on $(\mathbf{z}_t,\priors)$:
the cross term between its conditional mean and zero-mean residual vanishes.
Under the regularity conditions for flow matching, this field generates the
corresponding conditional marginal path \citep{lipman2023flowmatching}.
Individual linear training interpolants need not be its ODE trajectories;
the conditional expectation averages over different endpoint pairs.

The source in Eq.~\eqref{eq:conditional_source} has conditional density
$q_1(\mathbf{z}\mid\zc)=\mathcal{N}(\mathbf{z};\zc,\sigma_s^2\mathbf{I})$,
where the right-hand side is the Gaussian density evaluated at $\mathbf{z}$,
with mean $\zc$ and covariance $\sigma_s^2\mathbf{I}$. Its per-coordinate
variance is $\sigma_s^2$. This conditioning differs from conditioning on the visual
priors alone:
\begin{equation}
    q_1(\mathbf{z}\mid\priors)
    =\E_{\zc\mid\priors}\!\left[
      \mathcal{N}(\mathbf{z};\zc,\sigma_s^2\mathbf{I})\right].
    \label{eq:source_given_priors}
\end{equation}
The distinction above formalizes why the codec-centered source is not
redundant with the visual condition: $\priors$ need not identify the complete
bitstream or the exact codec state that centers the source. At the same time,
the conditional-flow interpretation should not be overextended. It does not
guarantee recovery of every paired clean latent or exact matching of the full
image posterior conditioned on a bitstream, and the fixed-construction
regression argument does not establish convergence when the codec and visual
interface are optimized jointly.

\subsection{Codec-Conditioned AYF-EMD Distillation}
\label{app:flow_identity}

We follow the AYF-EMD objective~\citep{sabour2025align} to distill the
Stage-I codec-conditioned flow into a one-step decoder, as summarized in
Eq.~\eqref{eq:stage2_total}. Teacher and student share the paired
interpolants $\mathbf{z}_t$ and visual priors $\priors$ associated with the
codec-centered source $\mathbf{z}_1=\zc+\sigma_s\boldsymbol\epsilon$.
The source-noise scale remains fixed throughout Stage~II; noise annealing
starts in Stage~III. We specialize the average-velocity parameterization
by fixing the destination time to $r=0$, so the student learns transport
to the clean endpoint and yields the one-step map in
Eq.~\eqref{eq:one_step_map} at $t=1$. The frozen Stage-I DiT serves as the
teacher and provides the student's initialization. The codec and all
condition-producing modules, including the evidence decoder, remain
frozen; only the student DiT is updated.

\subsection{Source Contraction and Its Scope}
\label{app:source_limit}

Stage~III contracts the stochastic source used for flow learning and
distillation onto the deterministic codec state used at deployment. The
following fixed-parameter analysis makes this source transition explicit.
Fix the codec, condition-producing networks, and generative-decoder
parameters, and let $d$ be the number of coordinates in the starting latent.
For $\rho>0$, the conditional source is
$q_{1,\rho}(\cdot\mid\zc)
=\mathcal{N}(\zc,\rho^2\sigma_s^2\mathbf{I})$; at $\rho=0$ it is the point
mass $\delta_{\zc}$, rather than a non-degenerate Gaussian density.
Here, $q_{1,\rho}$ denotes the conditional source law and $\delta_{\zc}$ the
point mass at $\zc$. Its covariance is $\rho^2\sigma_s^2\mathbf{I}$.
Writing $W_2$ for the 2-Wasserstein distance, every coupling with this point
mass has a constant second component, hence
\begin{equation}
    W_2^2\!\left(q_{1,\rho}(\cdot\mid\zc),\delta_{\zc}\right)
    =\E\|\rho\sigma_s\boldsymbol\epsilon\|_2^2
    =d\rho^2\sigma_s^2.
    \label{eq:source_wasserstein}
\end{equation}
Thus, at a fixed parameter snapshot, the source distribution approaches
the deployed codec state continuously in $W_2$ as $\rho\rightarrow0$. This
statement concerns the input distribution only; it neither requires nor
implies convergence in total variation to the point mass.

For a fixed condition, suppose the complete map
$\mathbf{z}\mapsto\mathcal{D}_g(\mathcal{G}_{\theta}(\mathbf{z};\priors))$
is $M$-Lipschitz in its starting latent, meaning its output distance is at
most $M$ times its input distance. Then
\begin{equation}
    \E\!\left\|
      \mathcal{D}_g(\mathcal{G}_{\theta}(\zc+\rho\sigma_s\boldsymbol\epsilon;\priors))
      -\mathcal{D}_g(\mathcal{G}_{\theta}(\zc;\priors))\right\|_2^2
    \leq M^2d\rho^2\sigma_s^2.
    \label{eq:source_output_bound}
\end{equation}
The bound compares perturbed and unperturbed outputs of the same fixed
decoder and therefore formalizes only the continuity of the deployment
transition; it is not a bound on reconstruction error to the source image.
During end-to-end training, both the source center and DiT parameters can
change, so the fixed-snapshot statements do not establish optimizer
convergence, monotonic distortion reduction, or preservation of the
teacher's output distribution. The deployed decoder is intentionally
deterministic and returns one reconstruction per bitstream rather than
sampling the complete conditional image posterior.

\subsection{Final-Image and Adversarial Objectives}
\label{app:final_losses}

Stage~III jointly optimizes the codec and one-step decoder for the final
rate--distortion--perception objective, both during source-noise annealing
and after the source becomes deterministic. The final-image term combines
complementary full-reference criteria:
\begin{align}
    \mathcal{L}_{\mathrm{img}}
    ={}&\E\!\bigl[
    \lambda_{\mathrm{mse}}\operatorname{MSE}(\xrec,\mathbf{x})
    +\lambda_{\mathrm{ms}}
      \bigl(1-\operatorname{MS\text{-}SSIM}(\xrec,\mathbf{x})\bigr)
    \nonumber\\
    &+\lambda_{\mathrm{lpips}}\operatorname{LPIPS}(\xrec,\mathbf{x})
    +\lambda_{\mathrm{dists}}\operatorname{DISTS}(\xrec,\mathbf{x})\bigr].
    \label{eq:stage3_image_loss}
\end{align}
The weights balance pixel accuracy with perceptual and structural similarity
\citep{wang2003multiscale,lpips,dists}.
As in the evidence objective, MSE averages squared differences over all
pixels and RGB channels. The expectation averages image pairs and the current
Stage-III source perturbation. The rate term is the estimated bpp defined in
\appref{app:codec_rate}.

For adversarial supervision, we adapt the vision-aided discriminator
formulation \citep{vfm_aid_gan_loss}. Let $\mathcal{J}$ be the selected set of
DINOv3 block indices for the discriminator, and let
$\operatorname{Head}_{\omega,j}$ be its trainable head at index
$j\in\mathcal{J}$. These indices need not be the semantic-conditioning
indices $i_{\ell}$. The discriminator at each level is
\begin{equation}
    D_{\omega,j}(\mathbf{x})
    =\operatorname{Head}_{\omega,j}\!\left(
      \Phi_{\mathrm{DINO}}^{(j)}(\mathbf{x})\right),
    \label{eq:dino_discriminator}
\end{equation}
where $\omega$ comprises the head parameters and the DINOv3 backbone remains
frozen. Let $\ell_D^{+}$, $\ell_D^{-}$, and $\ell_G$ denote the real-image,
fake-image, and generator classification losses, respectively. Then
\begin{align}
    \mathcal{L}_{D}^{\mathrm{DINO}}
    &=\E\!\left[\frac{1}{|\mathcal{J}|}\sum_{j\in\mathcal{J}}
       \left\{
       \begin{aligned}
       &\ell_D^{+}\bigl(D_{\omega,j}(\mathbf{x})\bigr)\\
       &+\ell_D^{-}\bigl(D_{\omega,j}(\sg(\xrec))\bigr)
       \end{aligned}
       \right\}\right],
       \label{eq:dino_disc_loss}\\
    \mathcal{L}_{\mathrm{adv}}^{\mathrm{DINO}}
    &=\E\!\left[\frac{1}{|\mathcal{J}|}\sum_{j\in\mathcal{J}}
       \ell_G\bigl(D_{\omega,j}(\xrec)\bigr)\right].
       \label{eq:dino_gen_loss}
\end{align}
Here, $|\mathcal{J}|$ is the number of discriminator feature levels.
For the discriminator update, the reconstructed image is detached. For the
generator update, discriminator parameters are held fixed while gradients
through the discriminator and DINOv3 input reach the reconstructed image.
These distribution-level objectives differ from the paired evidence loss,
which directly compares each intermediate evidence image with its source.

\subsection{Training and Gradient Paths}
\label{app:training_paths}

Table~\ref{tab:training_paths} makes explicit how the progressive
adaptation stages differ in what they optimize. A frozen visual encoder can
still transmit gradients to its input in Stages~I and III. The generative
VAE decoder remains frozen throughout, but Stage~III image-loss gradients
pass through $\mathcal{D}_g$ to the upstream codec and DiT. Stage~II isolates
the flow-map distillation problem: the codec, state projection, evidence
decoder, and semantic projections are frozen, and detaching $\zc$ together
with the complete condition $\priors$ confines map-loss gradients to the
student. No evidence-loss update is applied in this stage.

\begin{table*}[!htbp]
\centering
\caption{\textbf{Optimization roles across the three adaptation stages.}
Stage~II isolates one-step flow-map distillation: only the student is
updated, using the frozen Stage-I teacher and fixed bitstream-derived
conditions.}
\label{tab:training_paths}
\small
\setlength{\tabcolsep}{4pt}
\renewcommand{\arraystretch}{1.13}
\begin{tabularx}{\linewidth}{Y l l l}
\toprule
Component & Stage I & Stage II & Stage III\\
\midrule
Codec and state projection & Train & Freeze & Train\\
Visual-evidence decoder & Train & Freeze & Train\\
Semantic projections & Train & Freeze & Train\\
DiT / average-velocity student & Train & Train & Train\\
DINOv3 and VAE encoders & Freeze & Freeze & Freeze\\
Generative VAE decoder & Not used & Not used & Freeze\\
DINOv3 discriminator heads & Not used & Not used & Train\\
\bottomrule
\end{tabularx}
\end{table*}

\section{Additional KID Evaluation}
\label{app:kid_evaluation}

\begin{figure*}[!htbp]
\centering
\includegraphics[width=\linewidth]{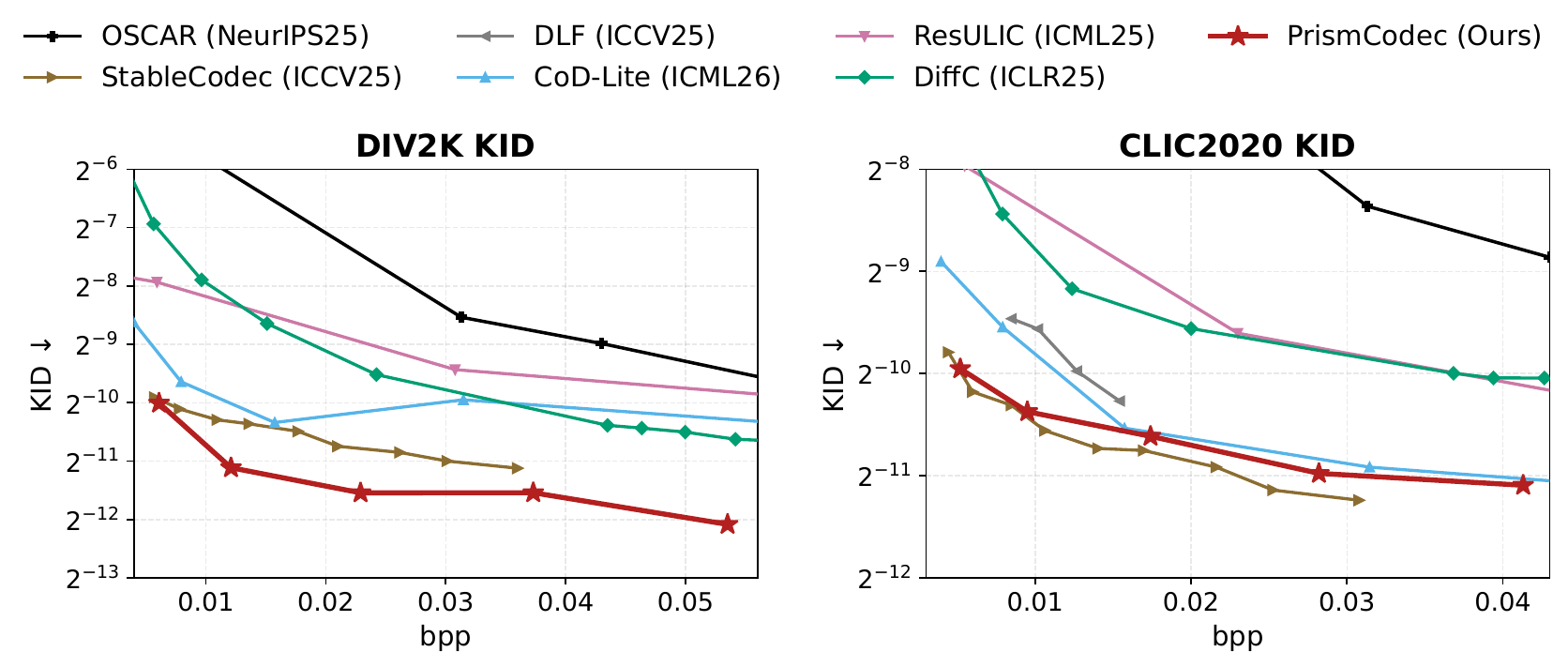}
\caption{\textbf{Rate--KID comparisons on DIV2K and CLIC2020.}
Markers denote measured operating points; lower KID is better.
The vertical axes use a base-2 logarithmic scale, and the bitrate windows
match those used in the main rate--perception comparisons.
\methodname{} is shown in red.}
\label{fig:kid_rd}
\end{figure*}

The main experiments use FID as a distribution-level complement to paired
fidelity metrics. We additionally report Kernel Inception Distance (KID)
\citep{kid} on the DIV2K validation and CLIC2020 test sets to provide a
second view of distributional realism. For \methodname{}, KID is computed
from the same $256\times256$ patches used for FID in
Sec.~\ref{sec:experimental_setup}, so the comparison changes the
distributional metric rather than the evaluation samples.
Fig.~\ref{fig:kid_rd} reports the measured rate--KID curves.

\section{Limitations and Future Work}
\label{app:limitations}

\methodname prioritizes rate--perception performance and effective adaptation
of a pretrained generative prior, rather than minimizing model size. As a
result, the current decoder remains relatively large compared with lightweight
compression models. Moreover, high-resolution images are decoded with
overlapping latent tiles to keep memory consumption manageable. These
computational costs accompany the substantial rate--perception gains obtained
by the current design, but leave clear opportunities for improving deployment
efficiency. Future work will explore more compact generative decoders and
architectures capable of processing full-resolution images directly without
tiled inference, while preserving the benefits of bitstream-grounded
one-step decoding.
\section{Additional Qualitative Comparisons}
\label{app:qualitative}

\begin{figure*}[!htbp]
\centering
\includegraphics[width=\linewidth]{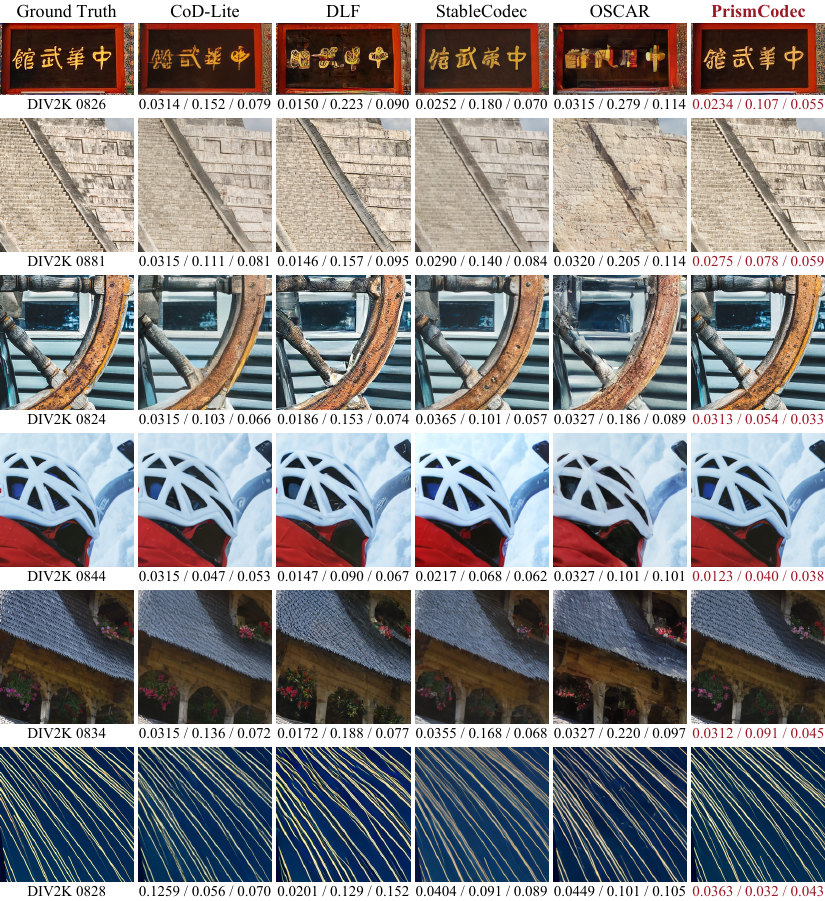}
\caption{\textbf{Additional qualitative comparisons on DIV2K.}
Rows show 0826, 0881, 0824, 0844, 0834, and 0828. Each method uses the same native
crop. Labels give full-image bpp / LPIPS / DISTS; lower values are better.
These examples examine signage, stepped boundaries, wood grain and spoke geometry,
small manufactured structures, architectural ornament, and thin light
trails, respectively.}
\label{fig:qualitative_additional}
\end{figure*}

Fig.~\ref{fig:qualitative_additional} extends the main comparison with six
additional DIV2K examples.

\end{document}